\documentclass[letterpaper, 10pt, conference]{ieeeconf}  % Comment this line out if you need a4paper

\IEEEoverridecommandlockouts                              % This command is only needed if 
\usepackage{amsmath}
\usepackage{amssymb}
\usepackage[hidelinks]{hyperref}
\usepackage{color}
\usepackage{graphicx}
\usepackage{subcaption}
\usepackage{multirow}
\usepackage{booktabs}
\usepackage{derivative}
\usepackage{mathtools}
\usepackage{algorithm}
\usepackage{algpseudocode}

\usepackage[backend=biber,style=ieee,natbib=true,doi=false,isbn=false,url=false,mincitenames=1]{biblatex} 
\DeclareMathOperator*{\argmax}{arg\,max}
\DeclareMathOperator*{\argmin}{arg\,min}

\newcommand{\ph}[1]{{\textbf{#1}:}} % paragraph header

\newcommand{\fadhil}[1]{{\color{black}#1}}

\title{\LARGE \bf
Semantic Belief Behavior Graph: Enabling Autonomous \\ Robot Inspection in Unknown Environments
}

\author{Muhammad Fadhil Ginting$^{1}$, David D. Fan$^{2}$, Sung-Kyun Kim$^{2}$, \\ Mykel J. Kochenderfer$^{1}$, and Ali-akbar Agha-mohammadi$^{2}$
\thanks{$^{1}$Department of Aeronautics \& Astronautics, Stanford University, Stanford, CA, USA
        {\tt\small {\{ginting,  mykel\}}@stanford.edu}}%
\thanks{$^{2}$Field AI, Mission Viejo CA, USA
        {\tt\small {\{david, sung, ali\}}@fieldai.com}}%
}

\usepackage{flushend}

\begin{document}
\maketitle

%%%%%%%%%%%%%%%%%%%%%%%%%%%%%%%%%%%%%%%%%%%%%%%%%%%%%%%%%%%%%%%%%%%%%%%%%%%%%%%%
\begin{abstract}
This paper addresses the problem of autonomous robotic inspection in complex and unknown environments. 
This capability is crucial for efficient and precise inspections in various real-world scenarios, even when faced with perceptual uncertainty and lack of prior knowledge of the environment. 
Existing methods for real-world autonomous inspections typically rely on predefined targets and waypoints and often fail to adapt to dynamic or unknown settings. 
In this paper, we introduce the Semantic Belief Behavior Graph (SB2G) framework as a new approach to semantic-aware autonomous robot inspection. 
SB2G generates a control policy for the robot, using behavior nodes that encapsulate various semantic-based policies designed for inspecting different classes of objects. 
We design an active semantic search behavior to guide the robot in locating objects for inspection while reducing semantic information uncertainty. 
The edges in the SB2G encode transitions between these behaviors. 
We validate our approach through simulation and real-world urban inspections using a legged robotic platform. 
Our results show that SB2G enables a more efficient inspection policy, exhibiting performance comparable to human-operated inspections.
\end{abstract}

% % Problem motivation, why this is important
% % This work in one cool sentence
% Gist of Methods
%% Overview
%% key interesting components
% Results

%%%%%%%%%%%%%%%%%%%%%%%%%%%%%%%%%%%%%%%%%%%%%%%%%%%%%%%%%%%%%%%%%%%%%%%%%%%%%%%%

%%%%%%%%%%%%%%%%%%%%%%%%%%%%%%%%%%%%%%%%%%%%%%%%%%%%%%%%%%%%%%%%%%%%%%%%%%%%%%%%
    \section{Introduction}
%%%%%%%%%%%%%%%%%%%%%%%%%%%%%%%%%%%%%%%%%%%%%%%%%%%%%%%%%%%%%%%%%%%%%%%%%%%%%%%%
% What is the problem?
% Why is it interesting and important?
% Why is it hard? (E.g., why do naive approaches fail?)
% Why hasn't it been solved before? (Or, what's wrong with previous proposed solutions? How does mine differ?)
% What are the key components of my approach and results? Also include any specific limitations. 

% Pictures:
% \ph{Pictures}
% Real Robot
% Challenging environment
% Inspection
% Stair climbing
% Exploration
% FSM

\begin{figure}[t]
    \centering
    \includegraphics[width=0.48\textwidth]{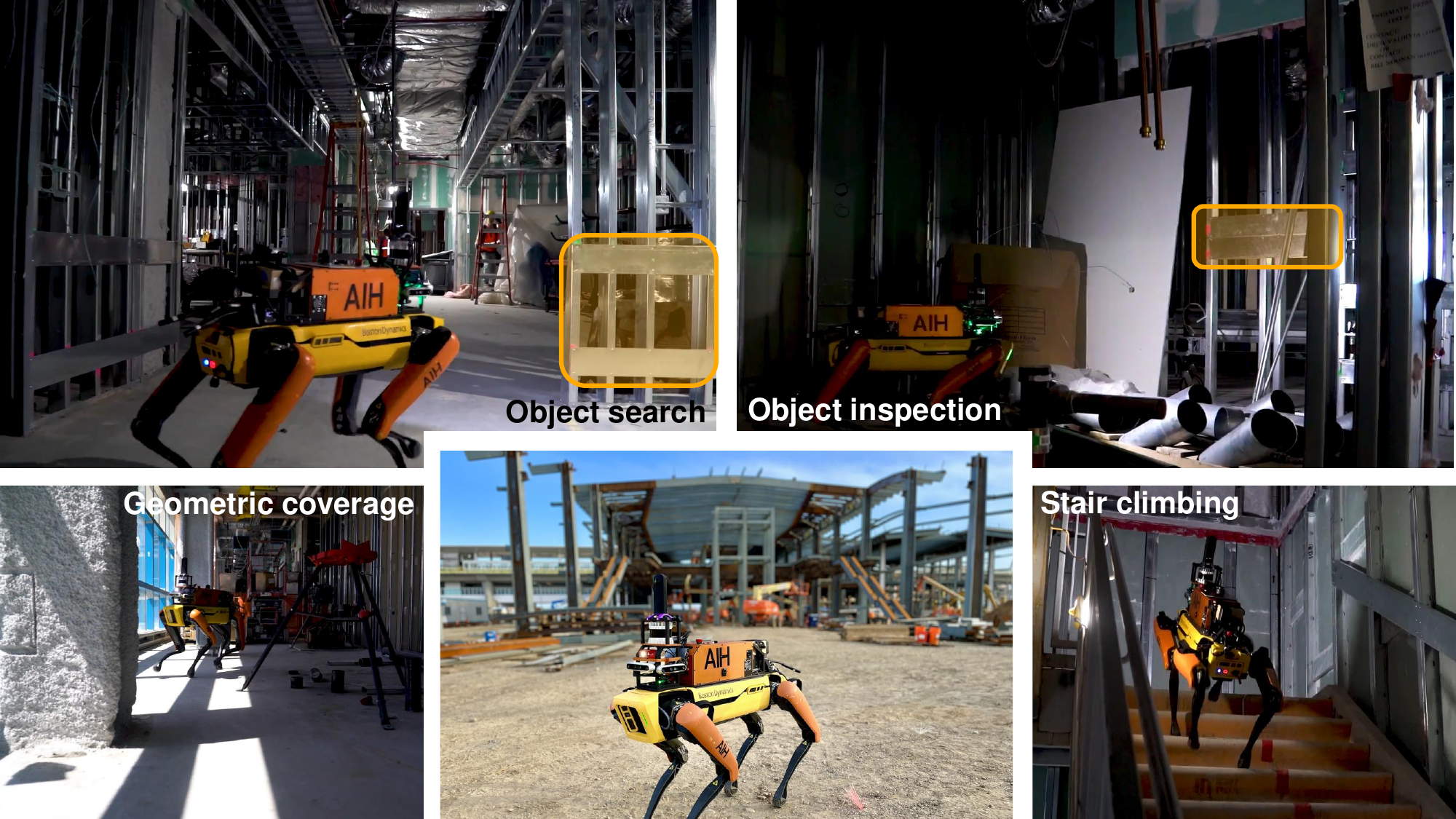}
    \caption{Autonomous robot inspection in urban environments. 
    This figure showcases various key semantic-aware behaviors performed by our robot to enable autonomous inspections.}
    \label{fig:cover_figure}
\end{figure}

% 1. What is the problem?
% \ph{What is the problem}
Consider a robot tasked with inspecting a complex and unknown environment autonomously. 
The robot needs to examine and interact with different types of objects dispersed throughout the environment, such as high-resolution object inspection, reading gauge measurements, and navigating through stairs and alongside humans (\autoref{fig:cover_figure}). 
To accomplish its objective, the robot must identify and reason about the semantic information of these objects and make decisions based on this information.
This capability for planning using semantic information is important to a wide range of real-world applications, including urban inspection~\cite{tan2021automatic,bouman2020autonomous,lattanzi2017review}, monitoring of oil and gas sites~\cite{gehring2021anymal}, exploration of subterranean environments~\cite{agha2021nebula, tranzatto2022cerberus, hudson2022heterogeneous, scherer2021resilient}, ocean exploration~\cite{khatib2016ocean, leonard2009autonomous}, and planetary exploration~\cite{balaram2018mars, agha2019robotic, touma2020mars}.
In this work, we address the problem of planning and decision-making using semantic information for autonomous robot inspection.

% 2. Why is it interesting and important?
% \ph{Why using semantic is important}
% Planning with semantic information offers many benefits for robots.
% Semantic information about objects and spaces enables robots to develop a contextual understanding of their surroundings beyond geometric cues~\cite{kostavelis2015semantic, kuipers2017bootstrap}.
% As a result, robot planning becomes more informed and adaptable, facilitating more intuitive navigation~\cite{pronobis2012largescale, bajracharya2009autonomous}.
% Moreover, semantic navigation establishes a bridge between human intentions and robot actions, as it aligns with human task specifications. 

% 

% 3. Why is it hard? (E.g., why do naive approaches fail?)
% \ph{Why planning with semantics is hard}
% 1. Presence of perception uncertainty: sensing range limitation, detection and localization error, occlusion
% 2. Environment uncertainty:Brittleness to changes in the environment such as object placement, limited or no prior knowledge about the environment
% 3. Folding semantic information into robotics planning and decision making that traditionally use geometric information.
However, performing semantic-based planning in unknown environments presents three main challenges, often rendering geometric-based planning methods ineffective.
The first challenge involves addressing perceptual uncertainty arising from semantic detection, which can be attributed to limitations in sensing range, false detections, localization errors, and occlusions.
The second challenge is in locating the semantic objects with the absence of or limited prior knowledge of the environment.
The third challenge is integrating semantic information into the robot's planning and control framework, which traditionally relies on geometric information.

% 4. Why hasn't it been solved before? (Or, 
% what's wrong with previous proposed solutions? 
% How does mine differ?)

% \ph{Why hasn't it been solved before?}
% Our work address these challenges for semantic based planning for autonomous robot navigation novel approaches
% - Account for these uncertainty,
% - design a semantic-aware behavior switching

% 5. What are the key components of my approach and results? Also include any specific limitations. 
% \ph{What are the key components of my approach and results?}
% \ph{Key components of the approach}
To address these challenges, we develop a framework called SB2G (Semantic Belief Behavior Graph) to enable robots to perform autonomous inspection tasks using semantic information.
To account for the perceptual uncertainty of semantic observations in planning, SB2G maintains both geometric and semantic information of objects of interest as a belief state.
The behavior nodes within SB2G represent different policies for controlling the robot in performing semantic-based inspection tasks.
To assist the robot in locating inspection targets with high confidence, we develop an active semantic search behavior that guides the robot in reducing belief uncertainty, directing it toward areas where it can gather more reliable semantic information. 
The SB2G edges govern the transitions between the behavior nodes, triggered either by attaining sufficient belief confidence for semantic-based behaviors or task specifications.

Using the SB2G framework, we demonstrate that the robot can autonomously perform inspections in real world environments without prior knowledge of the map or object locations.
This framework enables the robot to search for inspection targets and execute precise inspection behaviors efficiently.
Additionally, we compare the resulting robot behavior with the manual control exerted by humans during inspection tasks.

% \ph{Summary of results}

% In this work we propose an X framework (Semantic-aware Behavior Planning) smth
% \begin{enumerate}
%     \item Integrating semantic detection model
%     \item Semantic detection uncertainty quantification
%     \item Behavior design: normal point to point frontier/node geometric exploration, active search of the semantics, 
%     \item Mission design with behavior tree
% \end{enumerate}

% Results
% \begin{enumerate}
%     \item 
% \end{enumerate}

% Hypothesis:
% By reasoning semantic information, semantic detection uncertainty,
% we can plan a mission and behavior that
% are aware of semantic clue in the environment
% find and map semantic object in shorter amount of time, efficiently

% 6. Summary of contributions
% \ph{Contributions} 
% In this work, we propose an SB2G framework that enable semantic-based inspection tasks in unknown environment. 
In summary, our technical contributions are as follows:
\begin{enumerate}
    % \item We propose the SB2G framework, which uses uncertain semantic information to plan robot inspection behaviors in unknown environments. 
    \fadhil{
    \item We introduce the SB2G framework consisting of multiple object-dependent active semantic search types and semantic-based behaviors.
    % (\autoref{sec:problemformulation} and IV)
    \item We propose an active semantic search algorithm to observe hard-to-detect objects under semantic and perceptual uncertainty actively.
    % (\autoref{sec:behaviors})
    \item We design the behavior transition condition in belief space to enable reliable transitions between behaviors without needing a time-based transition condition.
    % (\autoref{sec:activesearch} and IV-D)
    \item We validate our framework through simulations and real-world demonstrations with a legged robot in various office buildings.
    }
    % \fadhil{, and analyze against human}. 
    % (\autoref{sec:experiments})
\end{enumerate}

% Figures:
% 1. First page figure
% 2. Problem formulation: Picture of robot (x), detected semantic(y), belief of y, task to do: inspect, climb, close open
% 3. SB2G figure: big behavior, small behavior
% 4. Transition figure: Abstract, example 
% opsional: SBG representation

\fadhil{
%%%%%%%%%%%%%%%%%%%%%%%%%%%%%%%%%%%%%%%%%%%%%
\section{Related Work}
\label{sec:relworks}
%%%%%%%%%%%%%%%%%%%%%%%%%%%%%%%%%%%%%%%%%%%%%
%%%%% 
\ph{Planning for autonomous inspection} 
% \begin{enumerate}
%     \item related to planning method in real-world inspection in unstructured environment
%     \item coverage planning to cover the area is a well studied topic
%     \item in many inspection scenario, only interested in searching and inspecting certain type of object
%     \item existing real-world inspection pre-defined view points or routes for inspection
%     \item More recent approach use different object search and mapping in the exploration strategy [1 Kostas, 2 Mihir, 3. Lu]
% \end{enumerate}
Our approach is related to planning methods that address the problem of autonomous robot exploration and inspection in real world settings. 
Coverage planning is a well-studied method to explore the entire area in the environment~\cite{bouman2022adaptive, peltzer2022fig, cao2021tare, charrow2015information, bircher2015structural}. 
This work addresses the problem where the robot only needs to search and inspect specific objects of interest in some part of the environment. 
Current state-of-the-practice approaches for object inspection usually rely on predefining routes and observation points or placing identifiable Apriltags or QR codes, making the process labor-intensive for humans~\cite{BostonDynamics2023GraphNav}. % TODO [4] seek
Recent works in the literature combine the object mapping method with coverage planning to tackle the object inspection problem~\cite{dang2018autonomous,lu2024semantics,dharmadhikari2023semantics}. % TODO cite Mihir, Lu, + 1 more

% n addition to map the whole
% environment, inspection tasks often require the robot to closely
% examine specific target objects of interest in some part of the
% environment [27]. Current state-of-the-practice approaches to
% inspecting specific targets usually rely on predefining routes
% and observation points for the robot or placing identifiable
% tags, such as Apriltags or QR codes, making the process labor-
% intensive for humans [4]. 

% \begin{enumerate}
%     \item Recent works active object mapping enables/crucial
%     \item % information gain, % cover the meshes % improving reconstruction quality
%     \item closely relevant to our work, Kostas
%     \item Mihir et al
%     \item Lu et al
%     \item learning-based approach, 
%     \item end with a gap: active search problem, false positive different class, semantic class uncertainty
% \end{enumerate}

\textbf{Semantic active mapping} methods have gained popularity for object-based search and mapping~\cite{asgharivaskasi2023semantic}. % cite object-centric exploration and mapping
The objective of active mapping is usually defined to maximize information gain~\cite{de2021real}, cover the meshes~\cite{dang2018autonomous}, and improve the reconstruction quality of the objects~\cite{papatheodorou2023finding}. 
Dang et al. propose a path planning algorithm that explores a new space while improving the object observation resolution on a volumetric map~\cite{grinvald2019volumetric}. % Cite
Similarly, Lu et al. sample candidate trajectories and add semantics-aware cost to improve the object-centric mapping~\cite{lu2024semantics}. % cite
More relevant to our work, Dharmadhikari et al. present three behaviors for volumetric exploration, semantic hole coverage, and object inspection, and switches between them~\cite{dharmadhikari2023semantics}.  % cite
Our work addresses a different semantic search and mapping problem. 
In our inspection scenario, the objects of interest are usually hard to detect without a careful search. 
The robot often receives false detection or detection with low object classification confidence. 
This problem requires a different active search method that accounts for the semantic detection uncertainties~\cite{atanasov2016localization, morilla2023perceptual}.

% \begin{enumerate}
%     \item We consider the task after actively search, switch
%     \item while there are ..
%     \item in general, determining the stopping condition
%     \item reactive condition over object detection is usually brittle 
%     \item switching condition for object inspection usually rely on 1, 2, 3
% \end{enumerate}
\textbf{Semantic-based task planning} is used to switch between object-based inspection behaviors. 
While some works have proposed semantic-based task planning frameworks in different application domains~\cite{wang2020home, galindo2008robot}, defining transition conditions for object search and mapping is nontrivial~\cite{placed2023survey}. %Cite active slam
Due to semantic detection uncertainty, the reactive transition between exploration and object inspection based on object detection signals can be brittle. 
Consequently, behavior transition for object inspection is usually based on a predefined time or assumes perfect semantic observation~\cite{dharmadhikari2023semantics,chen2023not}.

% performs autonomous exploration but also allows for small
% deviations from the exploration path in order to improve
% the resolution of observation of detected objects based on
% a volumetric map

% exploration path planning algorithm is
% designed such that in its every iteration it identifies a finite-depth
% collision free path that maximizes a gain related to exploring
% new space, as well as a gain related to the resolution of the
% observation of the previously mapped parts of the environment
% that relate to detected objects of interest.

\ph{Statement of contribution} With respect to the current literature, our work makes the following contributions. 
First, instead of combining exploration and object inspection behavior or only using a single object inspection behavior for all objects, our SB2G framework can facilitate multiple object search and inspection behaviors. 
Searching different objects (e.g., fire extinguishers, doors, stairs) requires different types of sensors and search behavior. 
Second, we develop a new active semantic search to confirm the object's presence with high confidence before transitioning to object inspection behavior. 
The active semantic search actively increases the confidence of the belief, making it possible to set the behavior transition condition based on the belief state. 
Finally, we extensively validate our framework in real world conditions on a legged robot in various office buildings. We highlight the importance of our SB2G framework for reliable semantic-based behaviors.
}
\section{Problem Formulation}
\label{sec:problemformulation}
%%%%%%%%%%%%%%%%%%%%%%%%%%%%%%%%%%%%%%%%%%%%%
% \ph{Overview} 
We first formulate the problem of semantic-based robot inspection in unknown environments.

\ph{Robot and object geo-semantic state} 
We define a robot state $x_k \in X$ and objects' geo-semantic state $y_k \in Y$ at time $k$. 
The robot state $x_k \coloneqq (x_k^p, x_k^q, x_k^a)$ consists of the position $x_k^p \in \mathbb{R}^3$, orientation $x_k^q \in SO(3)$, and internal state information $x_k^a$ such as robot's locomotion and sensor status. 
The semantic objects state $y_k$ represents the set of $N$ objects $y_k = \{y_{1,k}, \ldots, y_{N,k}\}$. 
Each object state $y_{i,k} \coloneqq (y_{i}^p, y_{i}^q, y_{i}^l, y_{i,k}^a)$ captures the object's position $y_{i}^p$, orientation $y_{i}^q$, class $y_{i}^l \in L$, and affordance status $y_{i,k}^a \in A$ of the object. 
We assume the objects' poses and classes are static, hence we omit the subscript $k$. 
In our example, the framework identifies and annotates objects of specific classes such as fire extinguishers, doors, and stairs, also indicating their status like `to be inspected' or `to be ascended'. 

\ph{Robot control and transition model}
Let $u_k \in U$ denote the control input, where the $d_U$-dimensional control space $U \subseteq \mathbb{R}^{d_U}$ accommodates various types of control inputs for both navigation and semantic inspection tasks. 
Examples of navigation control inputs include velocity commands and velocity limits, as well as robot locomotion modes \cite{sbg, brandao2019multi}. 
For semantic inspection tasks, control inputs may consist of actions like pitching the robot up or down to find objects, or activating sensing and data capture modules. The state evolution model $(x_{k+1}, y_{k+1}) = f(x_k, y_k, u_k, w_k)$ defines how the robot and object geo-semantic states evolve as functions of both robot control inputs and process noise $w_k$.

\ph{Geo-semantic observation variable and model}
When the robot operates in unknown environment, the state of the semantic objects $y_k$ is partially observable to the robot. 
If an object $y_{i,k}$ is visible from the current robot state $x_k$, the robot can obtain a geo-semantic observation $z_k \in Z$. 
The observation $z_k \coloneqq (z_k^p, z_k^q, z_k^l, z_k^s)$ consists of the object's measured position $z_k^p$ and orientation $z_k^q$, detected class $z_k^l$ and detection confidence score $z_k^s$~\cite{lei2022early}.
The observation model $z_k = h(x_k, y_k, v_k)$ encodes the relation between $(x_k, y_k)$ and $z_k$, where $v_k$ is the observation noise. 

\ph{Belief} A belief state $b_k \in B$ is a conditional probability distribution $b_k \coloneqq p(x_k, y_k \mid \mathcal{H}_k)$ over robot and objects' geo-semantic states given the history of observations and control inputs up to time $k$.
We use $b_k$ as the basis for decision-making. The belief is updated using a belief evolution model $b_{k+1} = \tau(b_k, u_k, z_{k+1})$, which we can be computed recursively:
\begin{align}
\label{eq:belieftransition}
b_{k+1} &= \alpha p(z_{k+1} \mid x_{k+1}, y_{k+1}) \times \notag \\
&\iint p(x_{k+1}, y_{k+1} \mid x_k, y_k, u_k) b_k dx_k dy_k, 
\end{align}
% \begin{align}
% \label{eq:belieftransition}
% b_{k+1} = \alpha p(z_{k+1} \mid x_{k+1}, y_{k+1}) \int_X p(x_{k+1} \mid x_k, u_k) b_k dx_k, 
% \end{align}
where $\alpha$ serves as a normalization constant.

\ph{Policy, reward, and cost}
Given the current belief $b_k$, the robot generates action according to a policy $u_k = \pi(b_k)$, with $\pi \in \Pi$. 
To find an optimal policy for the robot, we define semantic task rewards $r_l(b_k, u_k)$, $\forall l \in L$, and costs $c(b_k, u_k)$. 
\fadhil{The robot gets a reward when it successfully inspects objects, climbs stairs, or enters doors, while the costs include distance traveled by the robot or operation time.}

Given the preceding descriptions and formulations, we can formally define the problem:

\ph{Problem 1 (Semantic-based robotics inspection in unknown environment)}
Given a current belief $b_0$ and semantic object tasks defined by $r_l(b_k, u_k)$, find an optimal inspection policy $\pi$
\begin{align}
    \label{eq:problem_statement}
    \pi = &\argmax_{\pi \in \Pi} \mathbb{E} \bigg{[} \sum_{k=0}^\infty \sum_{l \in L} r_l(b_k, \pi(b_k)) - c(b_k, \pi(b_k))    \bigg{]}\\ \nonumber
    \text{s.t.~} &b_{k+1}=\tau(b_k, \pi(b_k), z_{k})\\\nonumber
    &z_k \sim p(z_k \mid x_k, y_k) ,\\\nonumber
    &x_{k+1} = f(x_k, y_k, \pi(b_k), w_k).
    % \\\nonumber
    % & r_l(b,\cdot) > 0, \forall b \in B^l,  r_l(b,\cdot) = 0, \forall b \notin B^l,
\end{align}
% Solving problem 1 POMDP problem is hard over the entire belief
% Sparse reward function

%%%%%%%%%%%%%%%%%%%%%%%%%%%%%%%%%%%%%%%%%%%%%
\section{Semantic Belief Behavior Graph (SB2G)}
\begin{figure}[t]
    \centering
    \includegraphics[width=0.49\textwidth]{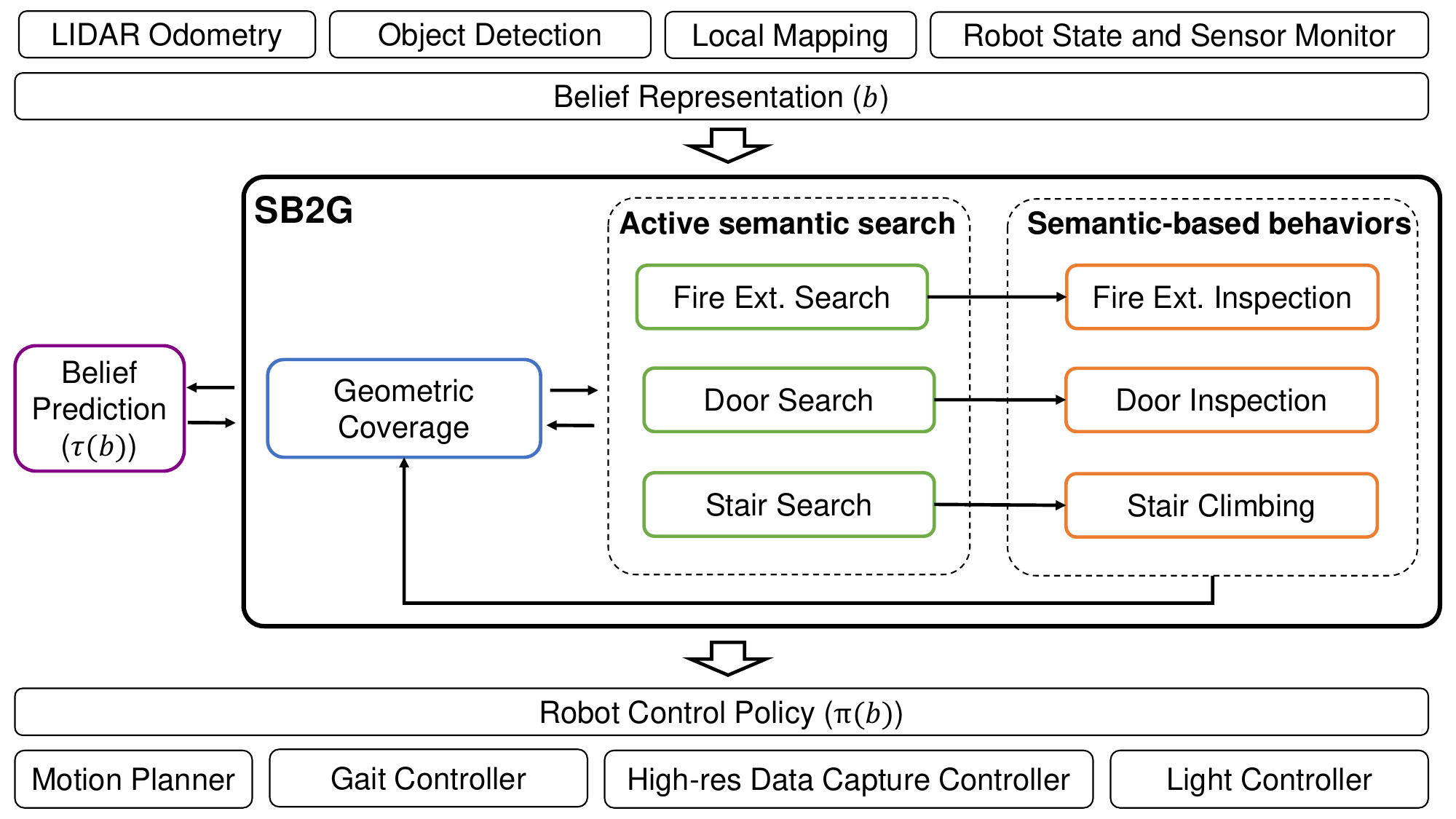}
    \caption{The system architecture of SB2G. SB2G gathers belief state $b$ from various perception modules. SB2G selects a behavior based on $b$ and uses belief prediction $\tau$ to compute a robot control policy $\pi$. The policy controls robot locomotion, high-resolution data capture, and lighting modules.
    }
    \label{fig:sb2g}
\end{figure}

% todo, make it more specific
\label{sec:sb2g}
%%%%%%%%%%%%%%%%%%%%%%%%%%%%%%%%%%%%%%%%%%%%%
% \ph{Motivating SB2G and Introduction}
To solve Problem 1, we propose the Semantic Belief Behavior Graph (SB2G) framework. 
In SB2G, nodes represent distinct robot behaviors, and edges denote transition conditions to switch between behaviors.
This framework is similar to finite state machines. 
SB2G enables the robot to efficiently navigate and inspect the environment by combining semantic understanding, geo-semantic state uncertainty, and behavioral decision making. 
The architecture of SB2G is shown in \autoref{fig:sb2g}. 

% \ph{Overview of the discussion}
In this section, 
we first detail the belief representation $b$ and prediction model $\tau$ that are used for decision making in SB2G (\autoref{sec:beliefprediction}). 
Then, we discuss three types of SB2G behavior for policy $\pi$ to control the robot: geometric coverage, semantic-based behaviors (\autoref{sec:behaviors}) and active semantic search behaviors (\autoref{sec:activesearch}). 
Active semantic search enables the robot to control the uncertainty of semantic beliefs for a reliable and informed transition to semantic-based behaviors.
Finally, we discuss how to design transition conditions to switch between behaviors and use SB2G for planning (\autoref{sec:transition}).

% We present Semantic Belief Behavior Graph (SB2G) framework.

% \ph{Assumption} 
% 1. Decompose objective and policy
% 2. Robot can localize the state without semantic state
% 3. Finite type of known objects

% \ph{Overview of SB2G}
% Figure 1 shows the graph representai

\subsection{Belief Representation and Prediction}
\label{sec:beliefprediction}
%%%%%%%%%%%%%%%%%%%%%%%%%%%%%%%%%%%%%%%%%%%%%
% We use recursive Bayesian estimation approach to estimate the belief of the robot $x$ and object geo-semantic state $y$ in the environment. 
% Estimating the belief requires a probabilistic model that quantify the likelihood of the semantic observation $z$.
% Later we use this model to predict future belief to compute policy $\pi$ for active semantic search behavior.
% We maintain belief representation for planning and develop the observation and belief likelihood for belief prediction in planning.

\ph{Belief representation}
We represent the robot belief and semantic object belief separately. 
The belief of robot pose $p(x^p, x^q)$ is represented as a Gaussian distribution. 
% $\hat{x}$
We use a LIDAR-based odometry method to estimate $p(x^p, x^q)$~\cite{reinke2022locus}. 
Meanwhile, we represent the semantic object belief $p(y)$ as an array of the belief of detected object $p(y_i)$. 
The object pose belief $p(y_i^p, y_i^q)$ is represented as a Gaussian distribution and estimated with an object-based localization using LIDAR and a color camera. 
The object class belief $p(y_i^l)$ is represented as a categorical distribution with $|L|$ categories. 
The class belief is estimated with a YOLO-based object detection~\cite{bochkovskiy2020yolov4}. 
Both the robot's other internal states \fadhil{(e.g., locomotion gaits, robot's sensor status)}, $x^a$, and the object's inspection status, $y^a$, are known to the robot.
% The other internal state of the robot $x^a$ and the status of the object $y^s$ is fully observable to the robot. 

\ph{Observation model}
The semantic observation of an object $y_i$ can be derived from a detection model and observation likelihood. 
We assume the robot can distinguish semantic observation $z$ from every object $y_i$. 
%Using observation model that capture data association from multiple objects is also possible~\cite{atanasov2016localization}.

% The detection model captures the probability of detecting object $y_i$ from robot $x$. 
The detection model quantifies the likelihood of the robot $x$ detecting an object $y_i$.
We model the detection probability for an object within the robot's field of view, $\mathrm{FoV}(x)$, using a decaying function based on distance
\begin{align}
p_d(y_i, x) := \begin{cases}
    p_{0,i} \exp\left\{ -\frac{|m_{0,i} - d(x, y_i)|}{v_{0,i}} \right\}& \text{if } y_i^p \in \text{FoV}(x) \\
    0& \text{otherwise},
\end{cases}
\end{align}
where $d(x, y_i)$ denotes distance between $x$ and $y_i$. 
The constants $p_{0,i}$, $m_{0,i}$, and $v_{0,i}$ specify the base detection probability, optimal detection distance, and the decay rate.

% \begin{align}
%     s_{stair, k} &= [\mathrm{trans}(T_{w, stair, k}), \mathrm{yaw}(T_{w, stair, k} - p_{\theta,k})] + v_k, \\ \nonumber
%     v_k &\sim \mathcal{N}(\textbf{0}, \textbf{\mathrm{R}}_{v,k}).
% \end{align}
% When the robot detect an object, the observation likelihood model the probability of the observation $z$. 
When the robot detects an object, the observation likelihood models the probability of obtaining the observation $z$.
Since $z$ is conditioned on the object state $y_i$, the observation measurements can be assumed to be independent of each other. 
We model the pose measurement likelihood $p_{pq}(z^p,z^q \mid y_i,x)$ as a Gaussian distribution with mean $(y_i^p, y_i^q)$ and covariance $(\Sigma_i^p, \Sigma_i^q)$. 
From experiments, we observe the noise of pose measurement and detection score increases with the distance to object $d(x,y_i)$ and true bearing angle $\beta(y_i,x)$. The noise also depends on the object class $y_i^l$. We model $(\Sigma_i^p, \Sigma_i^q)$ as
\begin{align} \label{eq:posecovariance}
    \Sigma_i^p = \mathrm{diag}( &(\sigma_{p,d} d(x,y_i) + \sigma_{p,\beta} \beta(y_i,x) + \sigma_{p,l} \gamma(y_i^l))^2, \\ \nonumber
    \Sigma_i^q = \mathrm{diag}( &(\sigma_{q,d} d(x,y_i) + \sigma_{q,\beta} \beta(y_i,x) + \sigma_{q,l} \gamma(y_i^l))^2. \nonumber
\end{align}
The constant $\sigma$ and function $\gamma$ are learned from data.

The measurement of semantic class $z^l$ can be assumed to be independent of $x$ due to the scale and orientation invariance of the object detection module. 
The semantic class likelihood $p_{l}(z^l \mid y_i^l)$ is derived from the confusion matrix of the object detector and can be learned from data.

Meanwhile, experimentally we observe the score measurement $z^s$ decreases with $d(y_i, x)$ and we model it as
\begin{align}
p_{s}(z^s \mid y_i^l,x) = p_l \exp\left\{ -\frac{|m_l - d(x, y_i)|}{v_l} \right\}.
\end{align}
Finally, the observation likelihood can be defined as 
\begin{align}
    p_z(z \mid y_i,x) \coloneqq p_{pq}(z^p,z^q \mid y_i,x)p_{l}(z^l \mid y_i^l)p_{s}(z^s \mid y_i^l,x). 
\end{align}

\ph{State transition model}
For the transition model of the robot pose $p_x(x_{k+1}^p, x_{k+1}^q \mid x_k)$, We use a unicycle model that captures longitudinal and lateral velocities of the system, which are characteristics often found in legged robots.
We assume deterministic transitions for other state variables $x^a$, such as the robot's locomotion mode and the sensor in use.

\ph{Belief prediction model} 
Based on the state transition and observation models, we formulate the belief transition function $\tau$ according to Eq. (\ref{eq:belieftransition}).
% Object detection model $h()$
% Probability of detection. Exponential (do we need this?) 
% $z = \{z^x, z^y, z^heading, z^class, z^score\}$ \\
% $p(z^score) = N(|x-y|, sigma)$
% $p(z^class) = confusion matrix(class that we see)$
% $p(z^heading) = N(relative heading to robot average heading)$
% $p(z^x) = distance based$
% $p(z^y) = distance based$
% $p(z) = p(z^score) * p(z^class) * p(z^heading) * p(z^x) * p(z^y)$

\subsection{Semantic-based Behavior}
\label{sec:behaviors}
%%%%%%%%%%%%%%%%%%%%%%%%%%%%%%%%%%%%%%%%%%%%%
% \ph{Definition}
Semantic-based behaviors enable a robot to perform specific semantic tasks.
When a semantic-based behavior is triggered in SB2G, it executes control actions according to a policy $\pi^{i-l}$ to accomplish a specific task $i$ for a particular semantic class $l$. 
To gain rewards $r_l$ for the task, the robot needs to execute the behavior in a certain range of the belief state $B^l$. 
For example, $B^l$ can be defined as a set of beliefs when the estimated object location $p(y^p)$ is in close proximity to the robot with a low uncertainty and the semantic class confidence $p(y^l)$ is high.

% \ph{3 behaviors}
The design of a semantic-based behavior policy depends on the task. In this work, we use three different types of behavior to enable a complete autonomous inspection: object inspection, stair climbing, and geometric coverage.

\ph{Object inspection} This behavior enables the robot to perform a close-range inspection task. In the urban inspection scenario, we are interested in inspecting fire extinguishers and doors. 
This object inspection involves controlling the robot according to a policy $\pi^{\mathrm{inspect-FE}}$ or $\pi^{\mathrm{inspect-Door}}$.
% The policy is a sequence of predefined procedure to capture high-resolution measurement of the object. 
For fire extinguisher inspection, the robot needs to read the pressure gauge. 
For door inspection, the robot needs to check whether the door is closed or open and measure the dimension of the door.
The robot achieves its inspection objective when it successfully measure the correct object. 
% This require the robot to have a high certainty of the object presence before performing the inspection behavior.
Performing an incorrect inspection in the absence of the correct object will waste operational time.

\ph{Stair climbing} 
This behavior is crucial for autonomous inspection in multi-level environments. 
The stair climbing control policy $\pi^{\mathrm{climb-Stairs}}$ is different between mobility system~\cite{hutter2016anymal,mihankhah2009autonomous,bouman2020autonomous}. 
For legged robots, the robot first needs to estimate the pose of the stair in the proximity of the robot using dense point cloud data. 
After positioning the robot in front of the stairs, the robot needs to switch its gait locomotion policy to a stair climbing mode before traversing the stairs. 
Successful stair climbing expands the robot's inspection capabilities considerably. 
However, failed execution due to incorrect stair localization can be catastrophic. 

\ph{Geometric coverage} 
When the robot is not performing semantic-based behaviors and active semantic search, the robot explore the unknown environment using the geometric coverage behavior. 
We use a coverage planning algorithm $\pi^{\mathrm{coverage}}$ to explore the obstacle-free space in the environment~\cite{bouman2022adaptive}. 
The policy plans robot trajectory to sweep the free space with the sensor footprint.
Geometric coverage ensures the robot to cover the unknown environment efficiently while searching for inspection targets and stairs.

\subsection{Active Semantic Search Behavior}
\label{sec:activesearch}
%%%%%%%%%%%%%%%%%%%%%%%%%%%%%%%%%%%
% \ph{Motivation}
To execute the semantic-based behaviors successfully, the robot needs to have a low belief uncertainty about the object state. 
However, in real world operations, achieving high-confidence estimates of the object state is challenging due to perceptual uncertainties 
To address this challenge, we develop an active semantic search behavior to guide the robot to perform actions to increase the confidence of the belief.

% To successfully implement semantic-based behavior in an autonomous robot, it's crucial to minimize the uncertainty in object state estimation, especially in real-world operations where perceptual uncertainty is high. To address this challenge, we introduce an active semantic search behavior, aimed at guiding the robot to perform actions that increase the confidence of its semantic beliefs.

% \ph{Problem definition}
When active semantic search is triggered, the robot execute a policy $\rho^l$ to reduce the belief uncertainty of $y_i$ for the expected class $l$. 
The policy $\rho^l$ is parameterized by the object class $l$ to account for the varying strategies required for locating different types of objects. 

% \ph{Target Belief}
The policy $\rho^l$ drives the current belief $b$ to a set of target beliefs $B^{target}$. The belief target for $\rho^l$ is a union of two belief set, $B^{target} = B^{l} \cup B^{a}$. 
The first set $B^{l}$ represents beliefs with high confidence in the semantic object $y_i$. 
The second set $B^{a}$ represents the alternative outcome of the semantic search where $y_i^l$ is not $l$ with a high confidence. 

To drive the belief to $B^{target}$, we use the entropy $\mathbb{H}(y_{i,0:T}\mid z_{1:T}, b_0)$ of the target object $y_i$ conditioned over the future observations. 
The conditional entropy is an appropriate objective function because it quantifies the amount of information needed to describe the belief of the target object $y_i$ given the probabilistic value of 
semantic observation $z$. The active semantic search behavior solves the following problem:

% TODO figure active semantic search
\ph{Problem 2 (Active semantic search)}
Given a current belief $b_0$, the target object belief $p(y_i)$ with expected class $l$, compute a policy $\rho^l$: 
\begin{align}
    \label{eq:activesearch}
    \rho^l_{0:T}(p(y_i)) = &\argmin_{\rho_{0:T}} \mathbb{H}(y_{i,0:T} \mid z_{1:T}, b_0)\\ \nonumber
    \text{s.t.~} &b_{k+1}=\tau(b_k, \pi(b_k), z_{k})\\\nonumber
    &z_k \sim p(z_k \mid x_k, y_k) ,\\\nonumber
    &x_{k+1} = f(x_k, y_k, \pi(b_k), 0).
\end{align}
The policy $\rho^l$ is executed until the belief $b$ reaches $B^{target}$.

To reduce the computational cost of solving Problem~2 in real-time on the robot, we make two assumptions. 
First, we assume the transition model is deterministic ($w=0$) because the transition noise of the robot pose over a short horizon is small enough for the planning problem. 
Second, we employ a sparse sampling method to reduce the search branching factor of the robot's action space. Our approach narrows the search on identifying the next poses that are both situated in obstacle-free zones and contribute to entropy reduction.
% Second, we sample the robot's velocity action space and focus our search for the next poses that are in obstacle-free zones and reduce the entropy.
%$\mathbb{H}$.
% Second, we sample the action space of the robot velocity command $(\dot{x}^p_{k+1}, \dot{x}^q_{k+1})$ and only search the next robot pose $(x^p_{k+1}, x^q_{k+1})$ that reduce the observation uncertainty of the robot and on obstacle-free space.

We compute the active semantic search policy by performing a branch and bound search in a receding horizon manner~\cite{kochenderfer2022algorithms}. At every time step, an optimal policy $\rho^l_{0:T}(p(y_i))$ is solved and only $\rho^l_{0}$ is executed. After the belief is updated with the actual $z$, we compute a new $\rho^l$.

% %%
% todo figure

% Algorithm 2: Active Semantic Search Behavior
% sampling-based, online receding horizon.
% Optimize with MPC
% Until it reach the belief of target behavior or else

% \david{I see your comments here and I very much agree, this needs a detailed figure, algorithm, and explanation, and perhaps some benchmarking }

\subsection{Behavior Transition in SB2G}
\label{sec:transition}
%%%%%%%%%%%%%%%%%%%%%%%%%%%%%%%%%%%%%%%%%%%%%
% \begin{figure}[t]
%     \centering
%     \includegraphics[width=0.45\textwidth]{Figures/BehaviorTransition.jpg}
%     \caption{}
%     \label{fig:behavior_transition}
% \end{figure}

% \ph{Definition}
In SB2G, behavior transitions are represented by the graph edges $e \in E$. 
The trigger condition to transition between behaviors is governed by the SB2G transition policy 
$\pi^t : N \times B \to N$. 
Then, based on the selected behavior, the SB2G graph policy $\pi^g$ returns a control policy for the robot $\pi = \pi^g(n_k, b_k)$.
% The transition policy $\pi^t$ outputs an edge $e$ switching to a different behavior or a self-loop edge $\{(n_i, n_i)\}$ to stay in the same behavior node. 

% $\pi^{g}(n_i, b_k) = e_i$
\ph{Trigger condition}
There are four approaches to designing the trigger condition in SB2G. 
First, we consider the edges that transition from active semantic search behaviors to semantic-based behaviors $\pi^{i-l}$. 
The trigger condition for these edges occurs when the belief $b_k$ is in the set of beliefs $B^{l}$ that can ensure the robot performs the semantic-based behaviors successfully. 
Second, edges transitioning from active semantic search behaviors to geometric coverage are triggered when $b_k \in B^a$.
Next, we consider the edges that go from geometric coverage to active semantic search behaviors. 
% Here, the robot switches to active semantic search if it detects an uninspected object yiyi​ and the current belief bkbk​ falls within Bsearch−lBsearch−l.
Here, the robot switches to active semantic search if it detects an uninspected object $y_i$ and the current belief is in $B^{search-l}$.
The set $B^{search-l}$ consists of beliefs having low-confidence object probabilities $p(y_i)$, enabling the active search to effectively reduce the uncertainty. 
Finally, transitions from semantic-based behaviors to geometric coverage are triggered when the robot has successfully completed its semantic tasks.
% The transition policy $\pi^t$ based on the trigger condition can be computed offline and called during the planning.
% The belief about the object is also to need to be inside $B^{search_l}$.
% Second, the confidence of the object's pose estimate and the class probability need to be above a certain level $b_k \in B^{search}$. 
% The belief set to trigger a semantic search behavior is designed to reduce i the probability of false semantic detection b ($y_{i,k}^a = \mathrm{`to be inspected'}$) 

% we compare the expected utility between performing an active semantic search behavior leading s 
% $\mathbb{E} \bigg{[} \sum_{k=0}^\infty \sum_{l \in L} r_l(b_k, \pi(b_k)) - c(b_k, \pi(b_k))    \bigg{]}$. 

\ph{Semantic-based planning with SB2G} 
We summarize the process of performing robotic inspection using SB2G until a terminal condition $B^{terminal}$ (e.g., inspected $N$ objects) in Algorithm \autoref{alg:sb2g}. 
% An example of a termination condition is when the robot has successfully inspect $N$ objects in the environment.
% Given all the components in SB2G, we summarize how to perform robotics inspection with SB2G until a terminal condition $B^{terminal}$ in Algorithm \autoref{alg:sb2g}.

% Algorithm 2: planning with SB2G
% Initial belief, initial behavior
% Based on belief, choose transition (section D)
% Execute behavior policy (section B and C).
% Algorithm 1 recaps
% \begin{algorithm}
% \caption{Planning with SB2G}\label{alg:sb2g_planning}
% \begin{algorithmic}
% \end{algorithmic}
% \end{algorithm}

% input: initial belief b_o, SB2G = (N,E), termination condition B^{terminal}
% b_k = b_0
% n_k = 'geometric coverage'

% while b_k \notin B^{terminal}
%     select next behavior n_{k+1} = \pi^{trans}(n_k, b_k)
%     compute policy for robot \pi_k = \pi^g(n_{k+1}, b_k)
%     apply u_k = \pi_k(b_k) to the system
%     observe actual z_{k+1}
%     update belief b_{k+1} = \tau(b_k,u_k,z_{k+1})
\begin{algorithm}
    \caption{Planning with SB2G}
    \label{alg:sb2g}
    \begin{algorithmic}[1] % The number indicates number every line
        \Require $b_0$, $SB2G = (N,E)$, $\pi^t$, $\pi^g$, $B^{terminal}$
        \State $b_k \leftarrow b_0$
        \State $n_k \leftarrow \text{`geometric coverage'}$
        
        \While{$b_k \notin B^{terminal}$}
            \State Select a SB2G behavior $n_{k+1} \leftarrow \pi^{t}(n_k, b_k)$
            \State Compute a control policy $\pi_k \leftarrow \pi^g(n_{k+1}, b_k)$
            % \State $u_k \leftarrow \pi_k(b_k)$
            \State Apply the action $u_k = \pi_k(b_k)$ to the system
            \State Observe the actual observation $z_{k+1}$
            \State Update the belief $b_{k+1} \leftarrow \tau(b_k, u_k, z_{k+1})$
        \EndWhile
    \end{algorithmic}
\end{algorithm}

% E(c(b_0, \pi^c)) = 

% \subsection{} planning with this
% \subsection{} analysis 

% 1. Behaviors

% 2. Behavior transition controller
% Outcome, reach that behavior state or move to other behavior state
% $u = \rho(b_k, B_i, B_j)$

% Belief, confidence of the object $p(s^i_{type})_k$ at timestep k, distribution of the located object
% relative robot position to the target object $s_{location}-x_location$

% Action: sensory. activate close-range sensing, turn on lights
%         robot movement: get closer, face the direction of the object, pitch up or down

% Observation: confidence, position

% Objective: reach desired belief part of behavior state.
% By getting closer, increase the confidence

% $b_goal in target or coverage$
% $b_o, current$

% Example:
% To stair: yolo, stair pilot 
% To object: yolo
% To enter doorways?

% 3. Behavior switching
% Maximize mission objective
% Trigger

% In our scenario,
% inspect climb object first, Max number of inspected object, and identified stairs
% climb stairs
% geometric coverage

% mu output pi if it's inside belief of the behavior, and rho for the most promising observation that hasn't observed or climbed by the robot

% TODO Figure of 3 behavior state with ..

%%%%%%%%%%%%%%%%%%%%%%%%%%%%%%%%%%%%%%%%%%%%%
\section{Experimental Results}
\label{sec:experiments}
%%%%%%%%%%%%%%%%%%%%%%%%%%%%%%%%%%%%%%%%%%%%%
% \ph{Overview}
% We evaluate the proposed semantic-based planning with SB2G for inspection in simulation and real-world environments. 
We evaluate the SB2G framework for autonomous inspection in simulation and real-world environments.

% TODO table of SB2G parameters
% active search
% T = 8
% n_pose = 20
% transition
% geometric to active search 0.75
% active search 0.9
% distance to target 2 m
% std deviation 0.5

\subsection{Simulation Results}
\begin{figure}[t]
    \centering
    \includegraphics[width=0.48\textwidth]{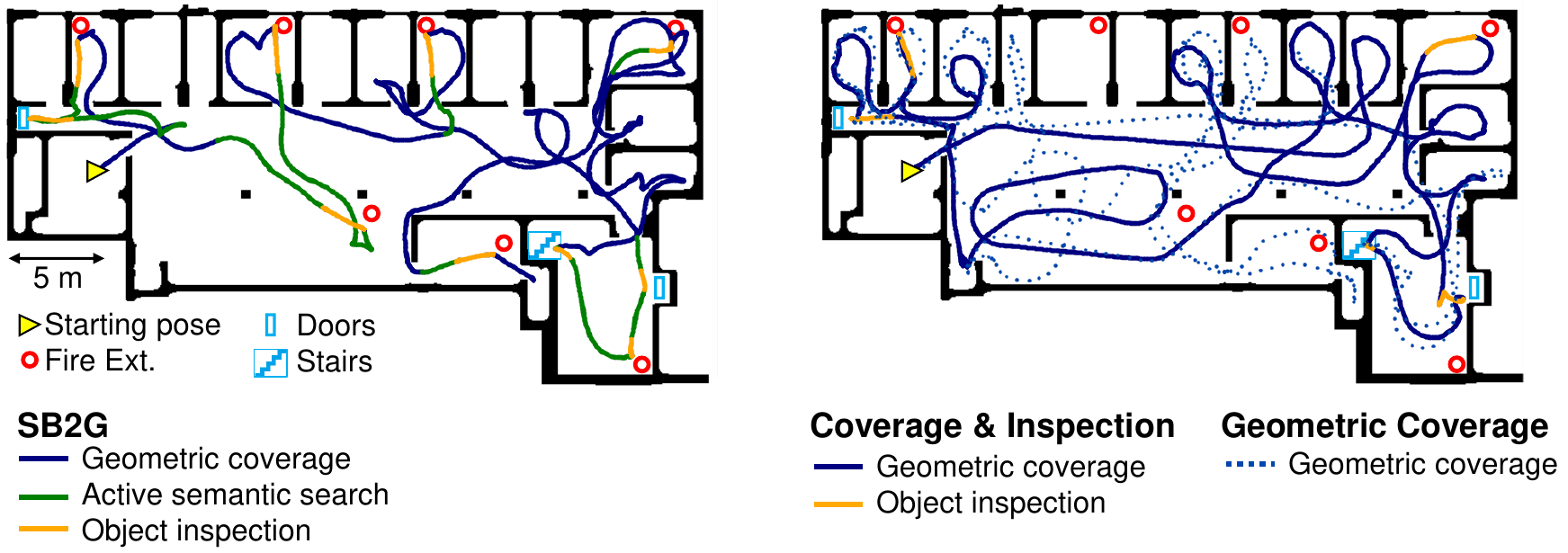}
    \caption{Comparison of robot paths for object inspections using SB2G and baseline methods in a simulation. Our method successfully locates and inspects all objects while following a shorter trajectory.}
    \label{fig:sim_path}
\end{figure}

% \ph{Experiment setup}
We perform simulations in a representative office environment with a size of $38 \times 20 \times 10 $ m. 
The simulation is carried out using the Gazebo simulator with a Boston Dynamics' Spot model. 
Ten semantic objects comprised of fire extinguisher, closed doors, and stairs are placed in the environment shown in \autoref{fig:sim_path}. 
The robot is tasked to perform object inspections.
The environment map and the true state of the objects are unknown to the robot.
The parameters used for the experiments are summarized in \autoref{tab:sb2g-parameters}.
The simulations were performed on a laptop with an Intel i9-11950H CPU.

\begin{figure}[t]
    \centering
    \includegraphics[width=0.48\textwidth]{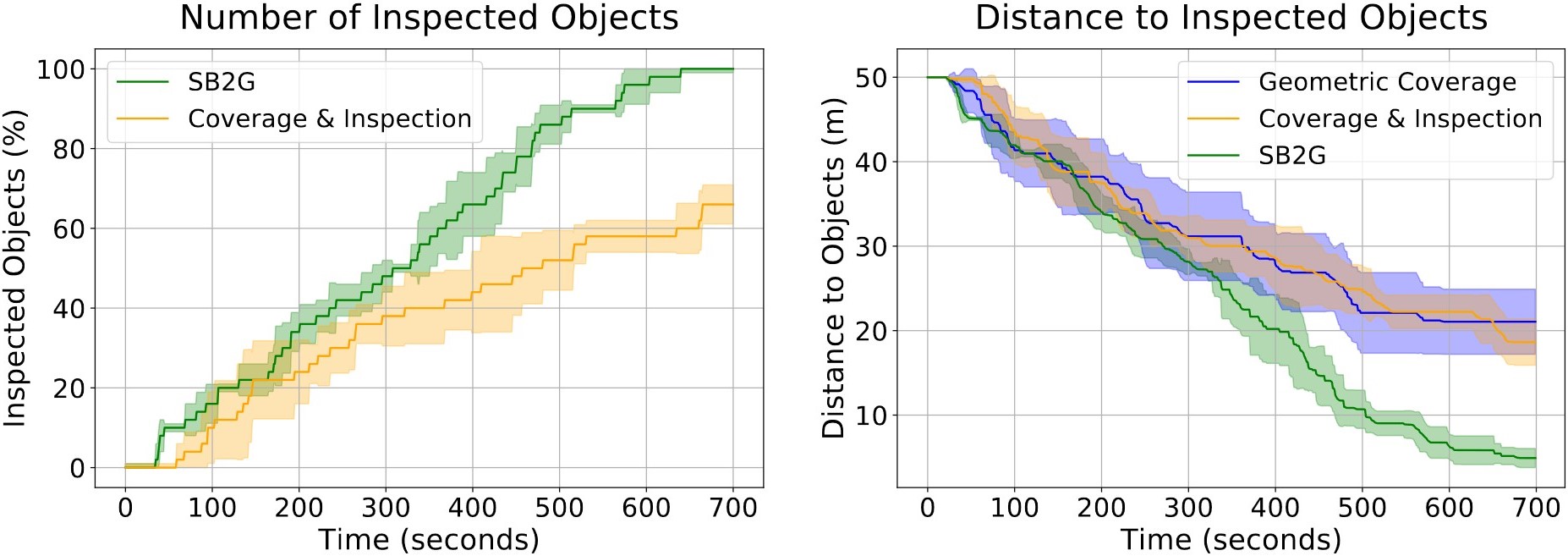}
    \caption{The number of inspected objects (left) and the sum of the closest distance to the all 10 objects over time (right). The closest distance to each object is initially set to $5$ m.
    % and the value is updated when the robot gets close to the object.
    }
    \label{fig:sim_quant}
\end{figure}

\begin{table}[t!]
\renewcommand{\arraystretch}{1.27}
\caption{SB2G parameters used in the experiments.}
\centering
\begin{tabular}{@{}ll@{}}
\toprule
\textbf{Parameter} & \textbf{Value} \\
\midrule
\textit{$B^{search-l}, \forall l \in L$} & $p(y_i^l) > 0.7$ and $\sum_i^p < 5 \, \text{m}$ \\
\textit{$B^{l}, \forall l \in L$} & $p(y_i^l) > 0.9$, $\sum_i^p < 1 \, \text{m}$, and $\mathbb{E}[d(x,y_i)] < 2.5 \, \text{m}$ \\
$T$ & $8 \, \text{s}$ \\
\bottomrule
\end{tabular}
\label{tab:sb2g-parameters}
\vspace{-5pt}
\end{table}

\begin{figure*}[h!]
  \centering
  \includegraphics[width=1.0\textwidth]{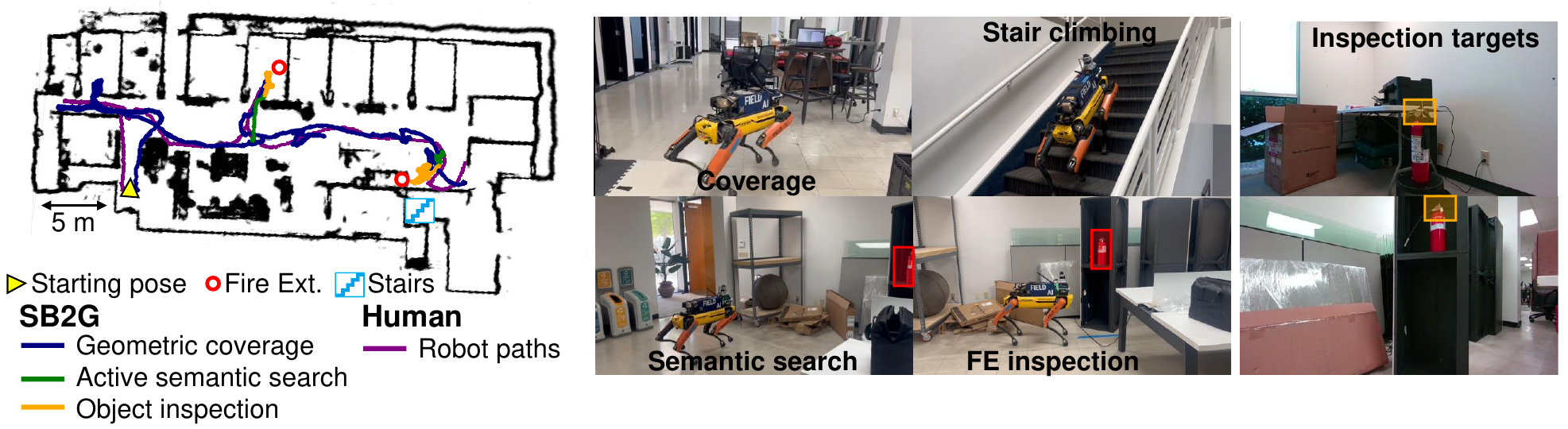}
    \caption{Experimental results of real-world autonomous inspections using SB2G. 
    The left figure compares the SB2G's robot paths with paths manually operated by a human during the search and inspection of fire extinguishers.
    The middle figures show our legged robot performing geometric coverage, semantic search, fire extinguisher inspection, and stair climbing. The right figures provide the camera view of the robot, demonstrating its capability to inspect the gauge of the fire extinguisher.}
  \label{fig:hw_results}
\end{figure*}

\begin{figure}[t]
    \centering
    \includegraphics[width=0.48\textwidth]{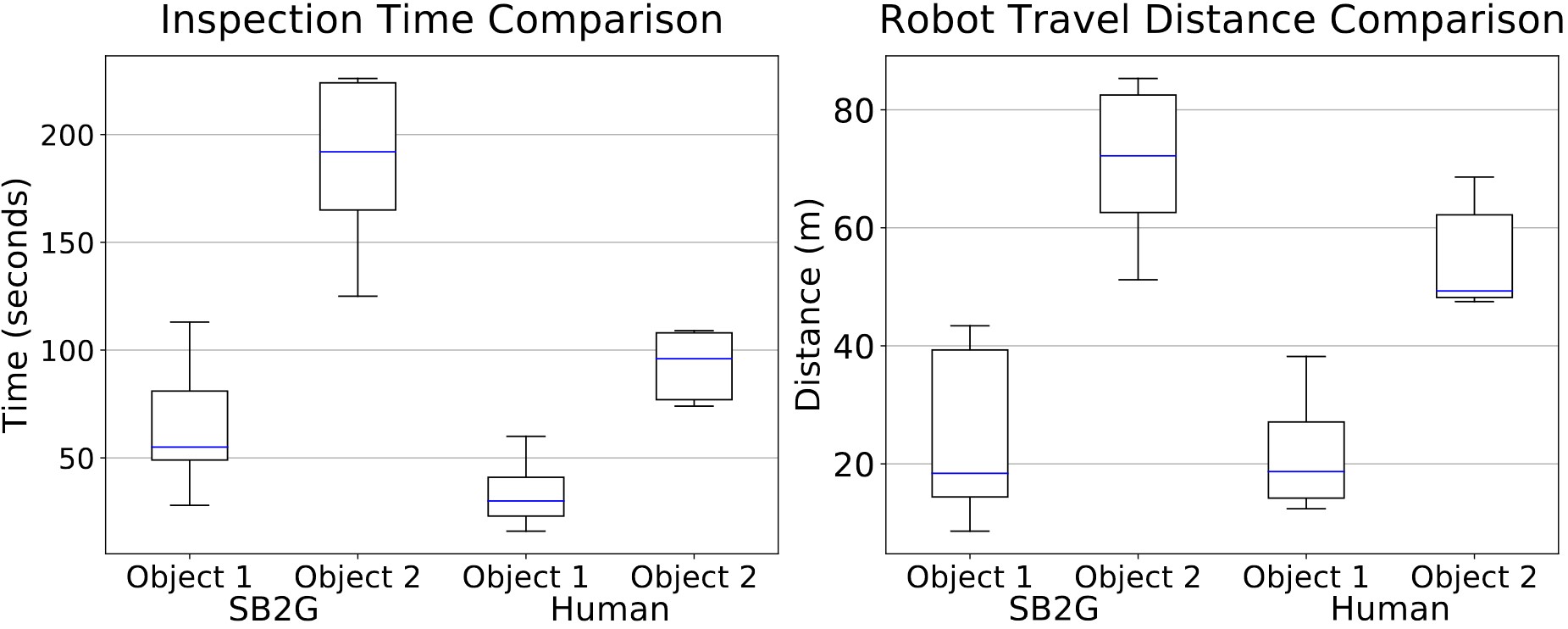}
    \caption{The comparison of inspection time and travel distance as the robot locates and inspects each object across 5 runs.}
    \label{fig:hw_plots}
    
\end{figure}

% \ph{Benchmark}
We compare our method to two methods:
\begin{enumerate}
    \item Geometric coverage: this method only perform coverage $\pi^{\mathrm{coverage}}$ to explore uncovered free space in the environment. 
    \fadhil{We compare our approach with a rollout-based coverage planning method that predicts future coverage using a sensor model similar to our sensor setup~\cite{bouman2022adaptive}.}
    \item Coverage and inspection: this method performs coverage $\pi^{\mathrm{coverage}}$ and object inspection ($\pi^{\mathrm{inspect-FE}}$, $\pi^{\mathrm{inspect-Door}}$, $\pi^{\mathrm{inspect-Stair}}$) without the active semantic search. The robot switches to object inspection behavior when the robot has high confidence in performing the inspection ($b \in B^l$).
\end{enumerate}
For each method, we run 5 simulation runs starting from the same location for $700$ s.

% \ph{Qualtitative Benefits}
% 0 Path of the robot, When to search, when to inspect. Example on one spot. 
% Highlight benefit of behavior transition, search for low confidence detection, 
% don't continue when false,
% Highlight benefit of active perception, pick the best path to observe things, find hidden object
\autoref{fig:sim_path} shows the comparison of the robot path and the behavior transition. 
The `coverage and inspection' method inspects fewer objects than SB2G.  This occurs because the robot rarely approaches the inspection targets closely and get reliable semantic observation for performing inspections. 
Meanwhile, SB2G generates a more efficient path by guiding the robot towards inspection targets by using the active semantic search. 
The active search policy $\rho^l$ controls the robot towards area where it can gather better semantic measurements.

% \ph{Performance}
% 1 Time/path length vsinspected object (percentage)
% 2 Time/path length vs distance object (m) 15 max
% (ours, geometric,)
% Table: method; number of detected; average distance to inspected object; pos error; (inspection rate(object/minute) optional) 

The inspection performance is presented in \autoref{fig:sim_quant}. 
The SB2G method outperforms the baseline methods considerably in the number of inspection over time. 
In all methods, the geometric coverage behavior helps the robot in rapidly approaching inspection targets, especially at the beginning of the run when semantic observations are not yet available.
However, without the active semantic search, the robot often fails to approach inspection targets closely enough, and consequently, is unable to execute inspection behaviors.
% While the naive inspection
% Highlight benefit of behavior transition, more efficient in searching
% Highlight benefit of active perception, actively reduce the belief uncertainty seen on the plot
%%
% Ideas for table
% Distance to object
% Confidence
% Localization error

% Entropy

%%%%%%%%%%%%%%%%%
\subsection{Field Test Results}
% Results:
% Photos of setup
% Qualitatively the path (Ours, human, geometric, naive). 2 plot
% Time or Path length vs Entropy of Detected object (Ours, human, naive)
% Table: Methods; Object found (3/3); time to find all object; Average path length

% \ph{Experiment setup}
We tested and deployed our solution in various office buildings in California, USA. 
In this paper, we focus on the results from experiments conducted in the Field AI office.
We conducted the experiment using a Boston Dynamics Spot robot. 
The robot is equipped with a LIDAR and 3 cameras for navigation and semantic observation. 
We use a YOLO-based model to detect fire extinguishers and a point cloud-based model for stair detection.
The detection model is not trained using the data gathered in the office. 
Two fire extinguishers are randomly placed in the environment. The robot does not have prior knowledge of either the office map or the objects' true locations.

% \ph{Benchmark with human} geometric + human. 5 times, 5 minutes or until 2 object were found
To evaluate the efficiency of our approach in real-world inspections, we compare our approach with the robot behavior manually operated by a human.  
The human operator also lacks prior knowledge of the object locations. 
We repeated the experiment five times, each time with different object locations. Each run lasts for 5 minutes or until two objects had been inspected.

% \ph{Qualitative vs geometric}
Using the SB2G framework, the robot can autonomously inspect fire extinguishers and climb the stairs (\autoref{fig:hw_results}). 
Despite numerous false-positive detection, our method effectively locates inspection targets by actively reducing semantic belief uncertainty. 
The robot accurately switches between different SB2G behaviors to accomplish a fully autonomous inspection task. 

% \ph{Result vs human}
The comparison between our SB2G approach and a human-controlled robot is summarized in \autoref{fig:hw_results} and \autoref{fig:hw_plots}. 
Qualitatively, as shown in \autoref{fig:hw_results}, our approach and the human operator produce similar inspection orders and paths. 
According to the statistics in \autoref{fig:hw_plots}, the human-controlled robot performs inspections more quickly and with less travel distance. 
The primary reason for this efficiency is that humans can more easily identify inspection targets, allowing for more direct paths and quicker inspections. 
Although the human operator lack prior knowledge of the office map, the intuitive understanding of potential room layouts allows the operator to search rooms efficiently for the objects. 
These findings can serve as guidelines for the development of more efficient and semantically aware behaviors.

% Qualitatively, 
% Qualitatively the path (Ours, human, geometric)
% Time or Path length vs Detected object (Ours, human, geometric)
% Style is the same
% Human better in speed and perception
% More efficient in joystick structurally, transitioning faster, inspection faster, false detection rejection
% Lessons learned: structurally search it efficient, 

%%%

% Comparison against:
% geometric coverage with semantic detection, 90\%
% coverage with semantic search with false positive
% Ours: 80\%
% Human

% Metric:
% Path of the robot, When to search, when to inspect, when to climb
% Detected object (confidence/accuracy). Entropy reduction, object reconstruction index (ORI)
% Inspected object. Number of inspected/measured object
% Path length vs inspected object or path length vs detected object
% Rate of inspection (inspected object vs time)
% Number of stops

%%%%%%%%%%%%%%%%%%%%%%%%%%%%%%%%%%%%%%%%%%%%%
\section{Conclusions}
\label{sec:conclusions}
%%%%%%%%%%%%%%%%%%%%%%%%%%%%%%%%%%%%%%%%%%%%%
% title
We introduced the SB2G framework that enables semantic-aware autonomous robotic inspection in uncertain and unknown environments. 
% Take away in method
SB2G uses semantic information to compute a control policy that guides robots through various inspection tasks. 
To enhance both the efficiency and accuracy of inspections, SB2G utilizes semantic belief uncertainty during planning and proactively mitigates this uncertainty before task execution. 
% Take away in results
Through simulations and real-world experiments, we showed that our approach achieves more efficient inspection behaviors and is comparable to human-performed inspections.
% generalizeable in method
% Our method is extensible to include a wide range of behaviours and generalizable in various environment domains.
% Future outlook
We believe this work represents an important step toward enabling more sophisticated semantic-aware behaviors in real-world inspection tasks.

% \ph{Future works}
% \begin{enumerate}
%     \item Semantic graph: Relation between objects
%     \item Contextual understanding of spaces: doorway
%     \item Optimization-based active perception 
% \end{enumerate}

%%%%%%%%%%%%%%%%%%%%%%%%%%%%%%%%%%%%%%%%%%%%
\vspace{-5pt}
% \section*{Acknowledgment}

% \bibliographystyle{IEEEtran}
% \bibliography{main}
\printbibliography
% Target: 50. 
% Application and motivating problem 15
% Related work 25: 10, 10, 10
% Our work or built upon 10

\end{document}